# Deep Learning-Driven State Correction: A Hybrid Architecture for Radar-Based Dynamic Occupancy Grid Mapping

Max Peter Ronecker[1,3], Xavier Diaz[1], Michael Karner[1,2] and Daniel Watzenig[2,3]

*Abstract*— This paper introduces a novel hybrid architecture that enhances radar-based Dynamic Occupancy Grid Mapping (DOGM) for autonomous vehicles, integrating deep learning for state-classification. Traditional radar-based DOGM often faces challenges in accurately distinguishing between static and dynamic objects. Our approach addresses this limitation by introducing a neural network-based DOGM state correction mechanism, designed as a semantic segmentation task, to refine the accuracy of the occupancy grid. Additionally a heuristic fusion approach is proposed which allows to enhance performance without compromising on safety. We extensively evaluate this hybrid architecture on the NuScenes Dataset, focusing on its ability to improve dynamic object detection as well grid quality. The results show clear improvements in the detection capabilities of dynamic objects, highlighting the effectiveness of the deep learning-enhanced state correction in radar-based DOGM.

## I. INTRODUCTION

Autonomous systems, ranging from self-driving vehicles to autonomous trains, are becoming integral components of modern transportation. At the core of such systems is the capacity to perceive and interpret the constantly changing surrounding environment. The accuracy and efficiency of this perception directly impact the decision-making processes of autonomous systems, thereby influencing their overall effectiveness and safety.

A fundamental perception algorithm is Dynamic Occupancy Grid Mapping (DOGM), which is capable of identifying static and dynamic elements as well as freespace in the environment. It extends classical occupancy grids [1] by incorporating a particle filter, enabling the estimation of dynamic objects inside the gridmap [3].

Typically, LIDAR serves as the primary sensor in these algorithms, chosen for its high-resolution point clouds and precision. However, LIDAR has a limitation in that it doesn't directly convey information about an object's movement, making Dynamic Occupancy Grid Mapping computationally expensive. This is because a substantial number of particles are needed to accurately estimate the dynamic states of objects. With advancements in radar technology, specifically in terms of accuracy and detection density, radar has emerged as a promising alternative for generating Dynamic Occupancy Grid Maps (DOGMap). In our recent work [4], [5], we introduced radar-based DOGM algorithms that effectively utilize radar's unique capabilities, such as range rate measurements, for optimized weight and state calculations. Although computationally lighter and with decent object detection capabilities, like many radar-based methods, these algorithms face challenges in distinguishing between static and dynamic objects as well as handling the noisy and sparse radar data. To address this limitation, we propose a novel hybrid architecture that integrates deep learning into DOGM to enhance state classification.

[1]SETLabs Research GmbH, 80687 Munich, Germany `first.last@setlabs.de`
[2] Virtual Vehicle Research GmbH, 8010 Graz, Austria `first.last@v2c2.at`
[3] Graz University of Technology, Institute of Automation and Control, 8010 Graz, Austria `first.last@tugraz.at`

Deep learning systems have shown impressive capabilities in various perception tasks, including image recognition [16], object detection [17], [18], and semantic segmentation [19]. They have also been successfully used for creating dynamic occupancy grids [24], [25]. However, most deep neural networks, while powerful, often lack interpretability and offer limited assurances in terms of safety. Conversely, traditional dynamic occupancy grid mapping offers a reliable base but typically falls short in achieving the same performance levels

**Contributions**

This paper introduces a deep learning-enhanced DOGM-state correction, integrated with radar-based DOGM, to create a hybrid algorithm. This approach is thoroughly evaluated on real-world data, demonstrating its efficacy:

1) We propose a neural network-based DOGM state correction, formulated as a semantic segmentation task. This network processes conventionally generated dynamic occupancy grid maps to correct falsely classified cells.
2) We present a DOGM architecture that merges conventional methods with deep learning based state correction, ensuring safety, robustness and performance.
3) The proposed hybrid architecture is extensively evaluated on real data and is able to significantly improve performance of radar-based Dynamic Occupancy Grid Mapping

## II. RELATED WORK

In the following an overview about Dynamic Occupancy Grid Mapping, both conventional and deep-learning based is given. We strictly focus on 2D occupancy grids and exclude research related to 3D Occupancy Grids as though similar in some aspects also bring their own set of algorithms and challenges.

### A. Conventional Dynamic Occupancy Grid Mapping

The original Bayesian Occupancy Filter (BOF) was first introduced in prior work [2]. This initial implementation utilized a four-dimensional grid, imposing substantial computational demands mainly due to the velocity computation. In response to these challenges, a more computationally efficient variant of BOF was presented in a subsequent work [3]. This enhanced approach integrated a grid representation with a particle filter, enabling simultaneous estimation of velocity and occupancy distribution within the grid. This has been further refined in [6], [7]. Moreover, the fusion of lidar and radar sensor measurements has been shown to enhance the filter's performance [8]. Additionally, a Dempster-Shafer representation of a Dynamic Occupancy Grid Map was introduced [9], [8], made computationally more efficient [10] and extended with object tracking and shape estimation. [11], [12], [13], [14].

However, the main focus of (dynamic) occupancy grids has been the use of lidar sensors as the primary information source, due to the dense point cloud data they provide. Up to this point, radar has primarily been utilized to support velocity estimation and accelerate convergence. Efforts have been made to adapt existing algorithms for radar-only use, such as [15] which use the approach of [8], [10] with radar data and integrates it with a clustering algorithm to provide object-level information.

Furthermore, based on [6], [7],modifications to state and weight calculations have been introduced in [4], [5], to better leverage the properties of radar data. However, it still struggles with noise of the radar sensor as well as dynamic-static separation.

## B. Deep Learning based Dynamic Occupancy Grid Mapping

Deep Learning has been used to improve several aspects of (dynamic) occupancy grid mapping. It has been applied to learn Inverse Sensor Models (ISM) for radar-based standard occupancy grid maps. In those cases lidar is used to generate the occupancy grid groundtruth [20], [21], can be extended to quantify uncertainty [21] or used in combination with a geometric ISM [22]. A novel self-supervised approach was introduced in [26]which predicts an accumulated future OGM. It has also been successfully shown to learn a multitask occupancy network, predicting freespace, occupancy and object detection at the same time [23]. Similarly, it has been shown that lidar based Dynamic Occupancy Grid Maps can be learned end-to-end [24] as well as in a multitask framework [25].

Although able to achieve good performance, completely replacing the probabilistic approach through a neural network creates new challenges with regards to reliability and interpretability of the system. For example it can be hard to guarantee that predicted freespace or occupancy is always correct. Furthermore, arguably the conventional DOGM approach already provides a solid foundation which can be enhanced wit deep learning rather than replaced. Hence, in this paper we propose an hybrid system in order to optimally use the advantages of both approaches.

## III. METHODOLOGY

This section details the used radar-based DOGM algorithm, the hybrid architecture as well as the neural work and training process used to learn the state correction.

### A. Hybrid Dynamic Occupancy Grid Mapping

The used DOGM algorithm is based on [4], [5] and it's main difference compared to other approaches is it pure range-rate based state computation. Opposed to using particles to determine if a cell is dynamic or static, the range rate is used. In some situations this can cause that cells are misclassified. Thus this approach is extended with a neural network which takes one or several of the calculated grid maps and predicts a DOGMap with corrected states. The predicted gridmap is then combined with the existing DOGMap using a heuristic based approach. In Fig. 1 an overview of the architecture is given.

*1) Radar-Centric Inverse Sensor Model:* In [5], we introduced an Inverse Sensor Model (ISM) tailored for radar data, advancing beyond traditional ISM approaches used in Dynamic Occupancy Grid Mapping. While conventional models primarily generate occupancy and free space information, our ISM simultaneously classifies each cell as dynamic or static based on the range rate. This approach suggests that cells proximal to a radar measurement with a nonzero range rate are more likely to be dynamic. Consequently, the measurement grid (MG) in our model assigns probabilities to each cell for being unknown, free, static, or dynamic, represented as:

$$P_{cell,meas} = \begin{bmatrix} P_{unknown} \\ P_{free} \\ P_{static} \\ P_{dynamic} \end{bmatrix} \quad (1)$$

This method enhances particle efficiency by generating particles exclusively in dynamic cells. However, it can misclassify slow-moving objects or vehicles moving perpendicularly to the sensor as static, potentially affecting tracking accuracy. To address these limitations, [5] suggests several non-machine learning techniques. In this paper, we replace this components with a deep learning-based state correction, to leverage the pattern recognition capabilities of neural networks.

*2) Dynamic Occupancy Grid Update:* In the Bayesian update process of a Dynamic Occupancy Grid Map, the latest measurement grid serves as the likelihood in our Bayesian formulation. Simultaneously, the pre-existing DOGMap is treated as the prior. This update step combines the new information (likelihood) with the existing knowledge (prior) to generate a more accurate and updated grid map.

$$P_{DOGM,upd.}(s|m) = \frac{P_{meas}(m|s) \times P_{DOGM}(s)}{P_{norm}(m)} \quad (2)$$

Where:
- $P_{DOGM,upd.}(s|m)$ denotes the updated probability of a cell's state **s** given the measurement grid **m**.

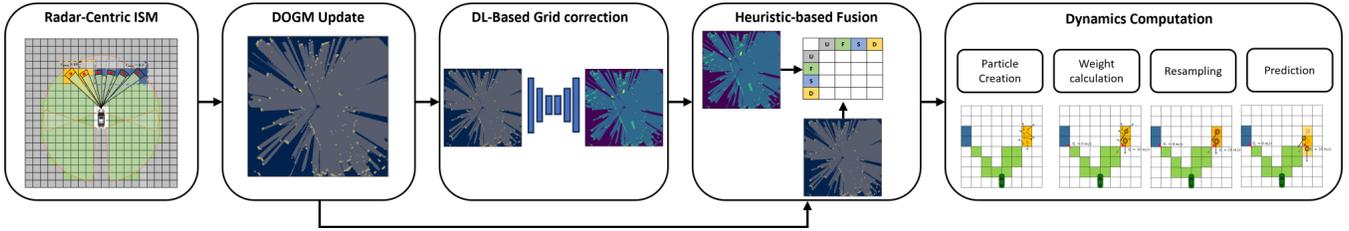

Fig. 1: **Hybrid Architecture**: It shows the general processing flow of the algorithm and how the neural network is integrated into the DOGM algorithm.

- $P_{meas}(m|s)$ is the likelihood which refers to the probabilities within the measurement grid for each state of a cell, based on the latest sensor data. For a specific state s, this likelihood is the probability value assigned in the measurement grid, reflecting how likely we are to observe based on the sensor data.
- $P_{DOGM}(s)$ is the existing probability of the cell states in the Dynamic Occupancy Grid Map, serving as the "prior" in Bayesian terms.
- $P_{norm}(m)$ is the normalizing constant, ensuring that the total probability sums to one.

*3) DL-based Grid Correction:* To address limitations in detecting dynamic objects caused by the radar-centric ISM, a grid state correction is essential. For instance, cells incorrectly labeled as static due to near-zero range rates are reclassified as dynamic. Furthermore, radar data noise and issues such as ambiguous line-of-sight geometry (e.g., multi-path reflections) may lead to erasing occupied areas or generating false positive occupancy. To address these issues, conventional methods have been proposed in [5], which are replaced with a neural network-based approach in this paper. This neural network is designed to generate a corrected Dynamic Object Grid Map. Detailed information about the network architecture and the training process can be found in Section III-B.

*4) Heuristic-based Fusion:* Ensuring consistent and predictable performance with Deep Neural Networks (DNNs) is still a challenge, but crucial for safety critical applications like autonomous vehicles and trains due to potential misses or false positives. Unlike DNNs, conventional Dynamic Occupancy Grid Mapping provides predictable outputs. To leverage the advantages of both systems, we propose a heuristic fusion approach. This method integrates the DNN outputs into the existing grid maps safely. The state combinations used in this approach are detailed in Table I. We adhere to the following rules to optimize performance without compromising safety:

- State changes are restricted to areas observed by the conventional DOGM.
- Any occupancy, whether dynamic or static, identified in the conventional DOGM must be preserved.
- State classification of occupancy can be changed if the DNN is sufficiently confident (output confidence score $c$).
- Occupancy is added to the grid only when a nearby measurement is within a specified distance threshold ($d_{meas} < d_{min}$), effectively reducing the risk of false positives.

*5) Dynamics Computation:* The particle filter is the final component of the DOGM Framework, employed for estimating cell dynamics and predicting movement. Its process involves particle creation, weight computation, resampling and prediction. Particles, representing position and velocity, are generated only in Dynamic Cells, which are subject to movement and thus require tracking. While a linear model is typically used for prediction, the framework allows integration of more advanced models. Weight computation based on the range rate, as detailed in [8] and also utilized in [5], helps in selecting the particles which are most probable to represent the tracked object. Resampling then filters out less likely candidates, maintaining only the promising ones. The final step is a prediction using a linear model to anticipate the dynamic object's future position. Additionally a transition matrix is applied to all states. We refer to [8], [7] and [5] for more details.

### B. Deep Learning based State Correction

As described in III-A.3 the overall goal of this component is the correction of falsely classified cell-states. To solve this with deep learning, we will formulate as a semantic segmentation task, using the existing DOGMaps as input and predicting a DOGM with corrected states. Different to [25], [23] which is based on sensor data or from a measurement grid [24], we use an already existing gridmap as input opposed to deriving a gridmap from the radar input. The reasoning behind is that the existing DOGM only needs minor corrections, hence this DNN works rather as a filter than a full grid computation.

*1) Architecture:* Our architecture is based on an Encoder-Decoder model with skip connections, similar to the one described in [19], [20], [23]. We have adapted this model to process the current and multiple previous gridmaps as input. They are concatenated and aligned in relation to ego-motion, similar to [24]. Through an ablation study, we found that using four gridmaps offers an optimal trade-off between performance and training time. It's noteworthy that although less maps decrease the system's performance it still yields reasonable performance. Therefore, the number can be varied to find a balance between performance and training duration. The output from our model is a corrected DOGMap, which maintains the same x/y dimensions as the input. It features four output channels that represent the states in a Dynamic Occupancy Grid Map. The standard configuration is a 100m x 100m grid with a resolution of 0.2m resulting in an input of 500x500x16 and a 500x500x4 output grid.

TABLE I: Heuristic-based Fusion state combinations

| | States | \multicolumn{4}{c}{DNN} | | | |
|---|---|---|---|---|---|
| | | U | F | S | D |
| DOGM | U | U | U | U | U |
| | F | F | F | S if $d_{meas} > d_{min}$ | D if $d_{meas} > d_{min}$ |
| | S | S | S | S | D if $c > c_{min}$ |
| | D | D | D | S if $c > c_{min}$ | D |

*2) Groundtruth Data:* For training, we utilize the NuScenes dataset, chosen for its extensive size and detailed ground truth annotations [27]. While the NuScenes dataset lacks a complete Occupancy Grid Map (OGM) ground truth, methodologies for generating them from LiDAR scans have been suggested in [20], [23]. However, for our specific application in Dynamic Occupancy Grid Mapping, we adopt a novel approach. We use bounding boxes, projected onto the 2D plane of the DOGMap, to correct the pre-existing DOGMap. These corrected DOGMaps then serve as ground truth during training. The correction process adheres to the following rules:

- Corrections are applied only to observed areas. Cells marked as unknown remain unchanged, even if they fall within a bounding box in the ground truth.
- The classification of elements as static or dynamic is based on their velocity in the ground truth data. Objects moving slower than 0.5m/s are classified as static, while faster objects are considered dynamic. The existing DOGM is updated to reflect these classifications.
- Free space within a bounding box is reclassified as occupied, with the classification (static or dynamic) determined by the object's velocity.

This method selectively corrects key elements in the DOGMap, focusing on relevant objects in the environment. It is simpler to label and more straightforward to implement and reproduce compared to existing techniques that derive an occupancy map e.g. from LiDAR data [23], [20]. The labeling process is also illustrated in Fig. 2 for further clarification.

*3) Training:* We approach this task as a semantic segmentation problem, employing a Cross-Entropy loss function coupled with weight balancing. Weight balancing is used due to the uneven distribution in the gridmap, where dynamic cells are much less common than areas such as free space. We adopt a DOGMap to DOGMap approach. This method uses the latest and several previous gridmaps to update the current Dynamic Occupancy Grid Map. It operates on the premise that most of the DOGMap is accurate, with only specific sections requiring modification. This approach increases training complexity due to the interdependence of input and output, comparable to a reinforcement learning scenario.

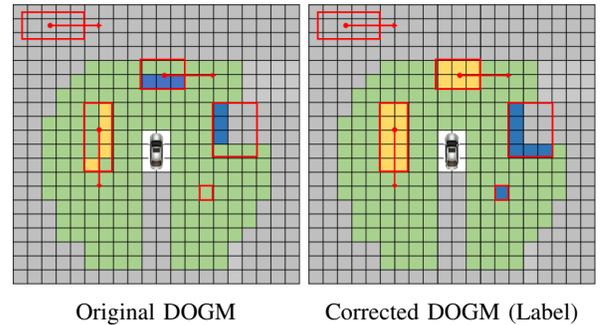

Fig. 2: **Visualization of the labeling procedure**: Bounding boxes are used to complete or modify the state in the observed area (gray=*unknown*,green=*free*,blue=*static* and yellow=*dynamic*).

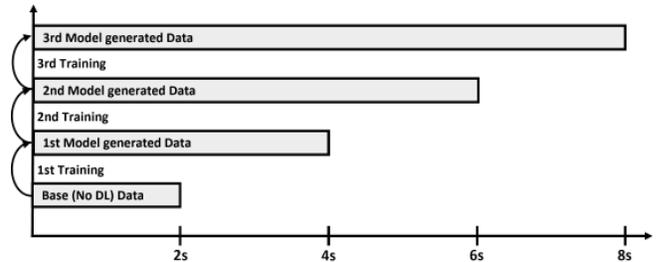

Fig. 3: **Training procedure**: The model undergoes incremental training, where it alternates between data generation and subsequent retraining (shown up to 8s)

Standard gradient descent optimization assumes independent input and output, which is not the case here, as the corrected DOGMap is feed back into the system. Training the network in a loop with the DOGM algorithm would be too slow, due to the time-consuming nature of accurate ground truth generation and the DOGM algorithm itself. Therefore, we adopt an incremental learning strategy similar to experience replay in (deep) reinforcement learning [30].

This incremental approach involves dividing scenes into 2-second segments for initial training. We then progressively increase the duration up to 20 seconds (maximal length NuScenes scene). This method ensures that the training data remains similar to the previously used distribution, mitigating the issue of later parts of a scene being too altered by previous network modifications. A detailed illustration of this training procedure is be provided in Fig. 3.

## IV. EVALUATION

To evaluate our method's effectiveness, we compare the deep-learning enhanced hybrid Dynamic Occupancy Grid Map algorithm with our non-machine learning radar-centric DOGM [5] as baseline and other radar-based object detection systems [33], [34] tested on the NuScenes Validation Set. This comparison focuses on two key aspects: the capability to detect dynamic objects and the overall quality of the occupancy grid.

## A. Evaluation Data & Metrics

The evaluation of our DOGM systems is subject to certain constraints. Firstly, the capability of DOGM is confined to tracking dynamic objects. For this, we utilize HDBSCAN [29] to cluster particles and then calculate their mean values for object-level information. A notable limitation of standard DOGM is its lack of bounding box outputs, a common feature in most object detection benchmarks. Moreover, the incorporation of the static environment, an integral part of the gridmap, is often overlooked in benchmarks, although recent advancements in 3D Occupancy Grids are beginning to address this issue. We incoporate those constrains in our evaluation.

For evaluation we use the validation split of the NuScenes Dataset [27]. NuScenes offers comprehensive ground truth data for both static and dynamic objects. However, its radar data, especially in terms of detection density, is not optimal, a limitation also recognized by other researchers [18], [28]. This causes a relative low overall detection performance but is still suitable to show the improvements of the method.

For comparison of the different approaches we use the following metrics.

- *Object Detection Metrics*: Our main metric is Average Precision (AP), commonly used for evaluating object detection performance, and calculate it as per the NuScenes Dataset guidelines. It involves averaging over multiple distance thresholds, which suits our approach given the absence of bounding boxes in our system. As confidence values we use a combination of particle weight and age as described in [5]. Additionally we use Recall (R) and Precision (P) to get a better understanding in which way the object detection is impacted.
- *Occupancy Grid Metrics*: Intersection-0ver-Union (IoU)is a frequently used metric in semantic segmentation evaluation, reflecting the similarity and overlap between two areas. Considering the parallels between semantic segmentation and our grid correction approach, we adopt IoU for assessing occupancy grid quality. To complement this we rely on Precision and Recall, which is useful in evaluating the Heuristic Fusion.

Unless specified differently, the Average Precision (AP) is calculated for dynamic objects moving faster than 0.5 m/s and inside a 100m x 100m grid centered around the ego-vehicle. For the minimum distance $d_{min}$ we take 0.5 and apply a max operator and $c_{min}$ larger then 0.5.

## B. Occupancy Grid Evaluation

This section evaluates the quality of the occupancy grid map, comparing $IoU^{Grid}$, Recall, and Precision metrics against the ground truth provided by a non-machine learning Dynamic Occupancy Grid Map (DOGM) algorithm. While free space metrics are included for completeness, they are less critical as they are based on the predefined field-of-view. The performance in detecting static occupancy remains largely unchanged, attributed to the minimal distance ($d_{min}$) threshold that prevents further occupancy accumulation. However, significant improvements are observed in detecting dynamic occupancy across all metrics, particularly in precision. This enhancement is credited to the deep learning based state correction's ability to track slow-moving objects, such as pedestrians, that would typically be misclassified as static. The beneficial effects of this capability on object detection are detailed in section IV-C.

TABLE II: NuScenes Grid Evaluation

| NuScenes (val) $IoU^{Grid}$ | | | | |
|---|---|---|---|---|
| Method | $IoU^{Grid}_{free}$ | $IoU^{Grid}_{static}$ | $IoU^{Grid}_{Dynamic}$ | $mIoU^{Grid}$ |
| Radar-Centric [5] | 98.0 | 73.9 | 17.4 | 63.1 |
| **Ours** | 97.7 | 71.8 | 20.1 | 63.2 |
| NuScenes (val) *Recall* (**R**) | | | | |
| Method | $R_{free}$ | $R_{static}$ | $R_{Dynamic}$ | Recall |
| Radar-Centric [5] | 100.0 | 74.2 | 17.4 | 63.9 |
| **Ours** | 100.0 | 72.0 | 20.4 | 64.1 |
| NuScenes (val) *Precision* (**P**) | | | | |
| Method | $P_{free}$ | $P_{static}$ | $P_{Dynamic}$ | Precision |
| Radar-Centric [5] | 98.1 | 99.1 | 81.6 | 92.9 |
| **Ours** | 97.7 | 99.6 | 88.0 | 95.1 |

Figure 4 illustrates enhancements in the grid. Notably, the applied heuristics allow us to maintain the conventional Field-of-View. Furthermore, we successfully identify several pedestrians (indicated by green boxes) previously undetected in the original gridmap, thanks to a relatively low $d_{min}$. However, large static and dynamic areas remain undetected despite being included in the predicted gridmap, due to insufficient surrounding measurements. This issue could be mitigated by increasing the minimum distance for measurements, but this approach risks generating numerous false positives and may not accurately depict smaller road users like pedestrians and bicycles. In our evaluation, we opted for a more conservative strategy, prioritizing the detection of pedestrians and bicycles due to their significance in radar perception challenges. Nonetheless, this parameter can be adjusted according to radar resolution, detection density, and target priorities.

Moreover, while the predicted gridmap often appears complete and aligns with the intended ground truth, directly modifying all grid states is impractical. Such changes would lead to many false positives and significantly deviate from the original grid, preventing accurate gridmap computation.

## C. Object Detection Evaluation

In this section, we focus on the object detection performance of the DOGM algorithm. In Table III, the results for dynamic object detection are presented. We evaluate five classes (car, truck, bicycle, motorcycle, and adult), which we consider the most relevant and also well-represented in the data. This evaluation includes all objects that are dynamic and inside the grid.

Our method significantly improves Average Precision across all categories, with notable gains in detecting bicycles, motorcycles, and adults. These categories are especially challenging for radar but essential for safe driving. A substantial increase in Recall demonstrates better recognition of relevant dynamic objects. Precision also shows improvement, though the extent is smaller relative to other metrics. The low Precision, at about 5%, can be partly attributed to the lower quality of radar data in the NuScenes dataset and the clustering algorithm, which may occasionally fragment objects. Despite this, our state correction technique effectively enhances the detection of dynamic objects while maintaining accuracy.

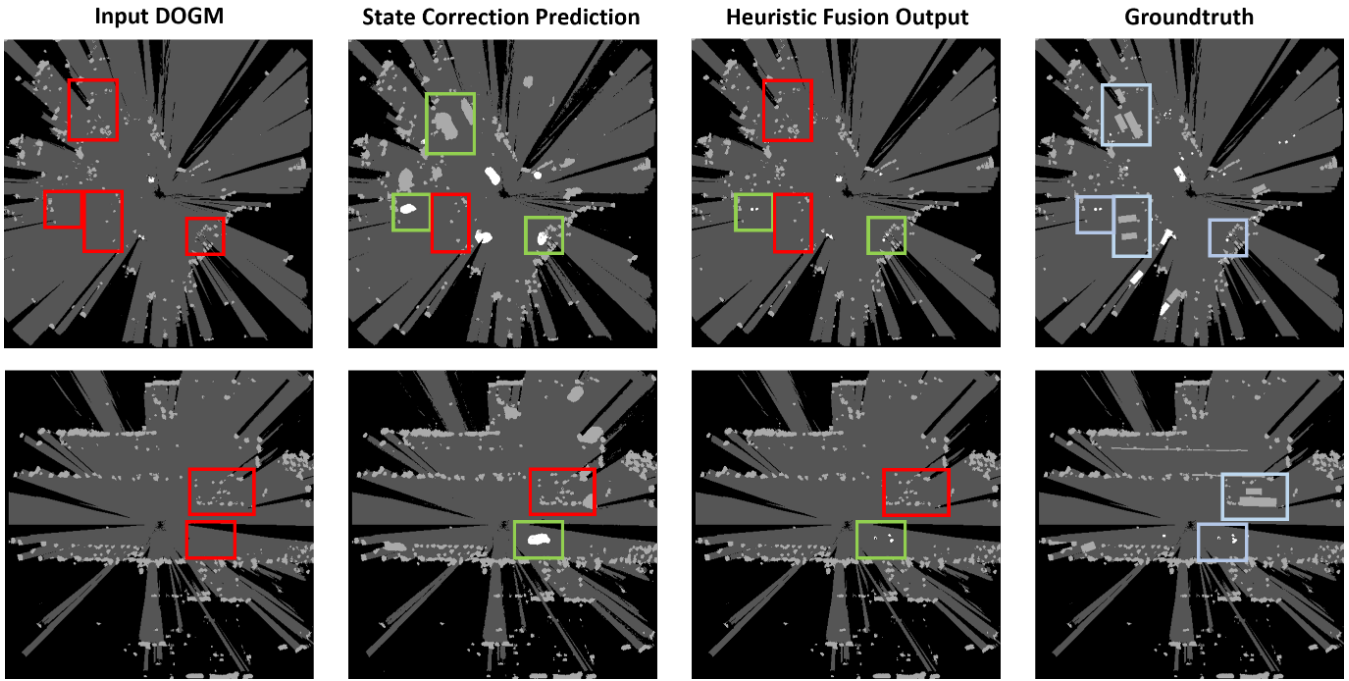

Fig. 4: **Grid comparison (two examples)**: The generated DOGMs as output of different components are shown. (black=*unknown*, darkgrey=*freespace*, lightgrey=*static*, white=*dynamic*). The green boxes indicate areas that are close to the groundtruth, red boxes mean that they are missing and blue highlights the respective area in the groundtruth.

Furthermore, we expand our analysis by comparing our system's performance against a range of advanced radar-based object detectors, as referenced in [33], [34]. This comparison, detailed in Table IV, focuses on the 'car' class and utilizes both AP (called mAP in [34]) and AP4.0 metrics as defined in [34]. Unlike other evaluations, this comparison includes objects regardless of their velocity, encompassing stationary objects as well, even though our Dynamic Occupancy Grid Map is only designed for tracking moving objects. We acknowledge that this comparison isn't perfect and disadvantageous for our algorithm, but can help in assessing the overall performance level of radar-based DOGMs. The results are shown in Table IV.

TABLE III: NuScenes Dynamic Object Detection (val)

| NuScenes (val) AP Dynamic | | | | | | |
|---|---|---|---|---|---|---|
| Method | $AP_{car}$ | $AP_{tr}$ | $AP_{Bc}$ | $AP_{Mc}$ | $AP_{adu}$ | mAP |
| Radar-Centric [5] | 19.3 | 14.2 | 1.2 | 13.6 | 0.1 | 9.7 |
| **Ours** | 27.1 | 18.4 | 8.9 | 28.6 | 4.9 | 17.6 |
| NuScenes (val) Recall (R) and Precision (P) | | | | | | |
| Method | $R_{car}$ | $R_{tr}$ | $R_{Bc}$ | $R_{Mc}$ | $R_{adu}$ | Recall |
| Radar-Centric [5] | 25.4 | 19.2 | 1.6 | 18.4 | 0.24 | 12.9 |
| **Ours** | 35.3 | 24.8 | 12.0 | 37.0 | 10.4 | 23.9 |
| Method | $P_{car}$ | $P_{tr}$ | $P_{Bc}$ | $P_{Mc}$ | $P_{adu}$ | Prec. |
| Radar-Centric [5] | 13.1 | 4.1 | 0.7 | 5.1 | 0.2 | 4.6 |
| **Ours** | 13.9 | 3.9 | 1.2 | 6.5 | 3.0 | 5.7 |

TABLE IV: NuScenes Object Detection Evaluation (val)

| NuScenes (val) Car | | |
|---|---|---|
| Method | $AP4.0_{car}$ | $AP_{car}$ |
| PointPillars[32], [34] | 38.70 | 22.44 |
| KPPillars[33] | 41.47 | 24.37 |
| KPBEV[34] | 42.27 | 25.26 |
| KPPillarsBEV [34] | 43.68 | 26.42 |
| **Ours** | 33.00 | 16.36 |

While our radar-based Dynamic Occupancy Grid Mapping doesn't match the performance of state-of-the-art object detectors, it still operates at an acceptable level. Importantly, its ability to include the static environment as part of its functionality offers potential to further increase object detection performance.

## V. LIMITATIONS AND FUTURE WORK

Our Hybrid DOGM algorithm has been tested using a single network architecture, specifically an Encoder-Decoder model. Exploring additional, more modern network architectures e.g. Transformers could potentially enhance performance. Moreover, the direct linkage between input and output (DOGM to DOGM) complicates the training process and makes hyperparameter optimization, such as adjusting the minimum measurement distance $d_{min}$, both time-intensive and challenging. Adopting an approach that transforms radar detections or measurement grids into a corrected DOGMap, essentially learning a DOGM Inverse-Sensor Model similarly to [24], [25], could streamline this process. Such an approach could be easily incorporated into our existing architecture, potentially boosting performance without sacrificing reliability.

## VI. CONCLUSIONS

In summary, this paper introduces a novel hybrid architecture that merges radar-based Dynamic Occupancy Grid Mapping with deep learning-driven state correction. This methodology enhances the performance of, dynamic-static classification, thereby improving dynamic object detection and the quality of occupancy grids. It offers a balanced solution, leveraging deep learning to boost performance without sacrificing the reliability inherent in conventional DOGM algorithms. In future work, besides the points mentioned we plan to advance radar-based DOGM by integrating it with Birds-Eye-View object detection, enabling more effective extraction of object-level details from both static and dynamic environments.

## ACKNOWLEDGMENT

This work has received funding from the German Federal Ministry for Economic Affairs and Climate Action (BMWK) under grant agreement 19I21039A.